\documentclass[letterpaper, 10 pt, conference]{ieeeconf}  %

\IEEEoverridecommandlockouts                              %

\usepackage{graphics} %
\usepackage{epsfig} %
\usepackage{mathptmx} %
\usepackage{times} %
\usepackage{amsmath} %
\usepackage{amssymb}  %
\usepackage{tabularx}
\usepackage{booktabs}
\usepackage{float}

\title{\LARGE \bf
XRoboToolKit-T: Teleoperation with High Stability and Precision with Tactile Sensing for Contact-rich Manipulation}

\author{Xiwen Dengxiong$^{1}$, Xueting Wang$^{1}$, Ke Jing$^{2}$, Rui Li$^{1}$, and Yunbo Zhang$^{3}$%
\thanks{$^{1}$Rochester Institute of Technology, Rochester, NY}
\thanks{$^{2}$TikTok Pico Lab, San Jose, CA}%
\thanks{$^{3}$The Hong Kong University of Science and Technology, Guangzhou, China}
}

\begin{document}

\maketitle
\thispagestyle{empty}
\pagestyle{empty}

\begin{abstract}
Collecting high-quality robot data for contact-rich manipulation tasks is essential for enabling robots to acquire real-world skills. However, existing data collection solutions often lack the capability to obtain stable and high-frequency tactile feedback, limiting their effectiveness in contact-rich manipulation scenarios.
In this work, we propose a versatile teleoperation system with tactile-driven assistance to enable high-frequency and stable contact-rich manipulation. The proposed XRoboToolKit-T teleoperation system incorporates a tactile-informed force control architecture, designed to ensure both stable and precise force control in contact-rich manipulation during teleoperation. The stabilizer haptic module rapidly analyzes the normal force distribution and infers pseudo shear force, enabling real-time tactile-based assistance during manipulation. The refiner haptic module integrates a vision-language-action model to predict and refine manipulation actions based on tactile sensing data and task descriptions.
We apply the proposed teleoperation system to challenging contact-rich manipulation tasks, including grasping a deformable rubber pipette for liquid transfer and inserting a medical syringe into a vascular training pad, to demonstrate the effectiveness of tactile-informed force control. Furthermore, the system achieves higher data collection efficiency and improved manipulation stability compared to state-of-the-art teleoperation without tactile assistance.

\end{abstract}

\section{Introduction}

In contact-rich robotic manipulation tasks, especially those involving flexible or deformable objects, visual information alone is insufficient. Instead, force feedback serves as a critical sensory modality, enabling robots to interact effectively with the physical world. While developing robust models for these tasks requires the effective and efficient collection of force-enriched manipulation data, utilizing current mainstream teleoperation methods remains a significant challenge, although recent research has \cite{hughes2020simple} explored the use of data gloves to reconstruct human contact interactions with the physical world. A teleoperation system translates a human operator’s inputs into control commands to drive the robot. For tasks such as squeezing pipette bulbs or inserting syringe needles, an abrupt fluctuation in gripper force caused by aggressive operation can easily cause the manipulation to fail or trigger the robot’s emergency stop. On the other hand, because human input is inherently imperfect, even small, unintentional hand movements by the operator can lead to excessive force application and task failure. Beyond just the magnitude, the precise location where the force is applied to the object also needs to be controlled in order to accomplish the task.

Recent studies have explored robot teleoperation systems based on XR headsets \cite{zhao2025xrobotoolkit, ze2025twist2, luo2025sonic, dengxiong2024self} or exoskeleton devices \cite{chi2025open} to capture joint rotations and end-effector motions for robot control. These systems have demonstrated promising performance in tasks involving interaction with rigid objects. However, such teleoperation systems are less effective when handling deformable objects or tasks that require more \underline{stable} and \underline{precise} control of contact forces.

\begin{figure}
  \centering
  \includegraphics[width=\linewidth]{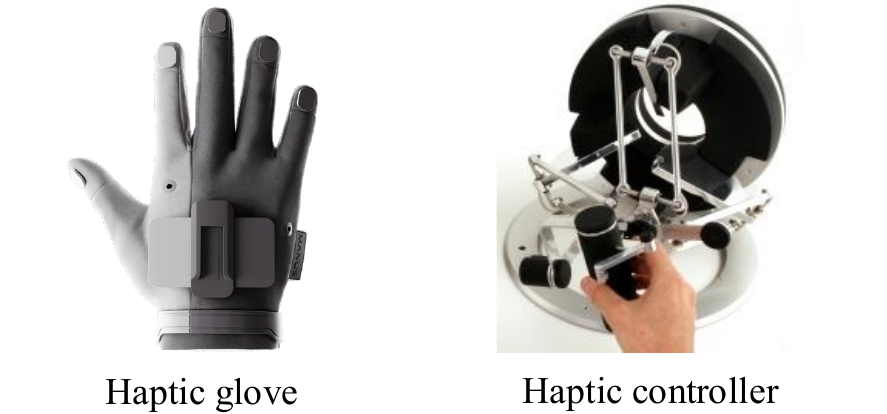}
  \caption{Current data collection solutions with force feedback.
  }
  \label{introduction}
\end{figure}

A feasible solution for contact-rich manipulation is to incorporate tactile information into the teleoperation system through either visual or mechanical feedback. However, simply visualizing tactile signals cannot provide enough information for the operator to understand and make adjustments, while it could increase the mental effort for the operator to process these signals. A data glove with force feedback (shown in Figure~\ref{introduction} left) is able to feed back the contact forces of the manipulation; however, it may not be quick enough to respond and save the manipulation from a failure even if the feedback is given. Moreover, the information provided by the data glove is usually not sufficient to enable a precise adjustment to the manipulation actions. The haptic teleoperation mechanism (shown in Figure~\ref{introduction} right) gives even less information compared to the data glove, and can only stop the robot's action if there is a large contact force. Therefore, tactile information needs to be utilized not only for feedback to the operator, but also for the control of the manipulation at an earlier stage. The remaining research question is: \textbf{how can we design a teleoperation system with tactile assistance that helps the operator stably and precisely control the force in contact-rich manipulation tasks}?

To this end, we propose XRoboToolkit-Tactile (XRoboToolkit-T or XRT-T), a tactile-informed teleoperation system designed to provide assistance for stable and precise force control for contact-rich manipulation. This system is built upon PICO's XRoboToolkit~\cite{zhao2025xrobotoolkit} and consists of two new modules: 1) a contact force \textit{stabilizer} and 2) a Vision-Language-Action (VLA) model-based action \textit{refiner}. Specifically, the contact force stabilizer analyzes the pressure distribution captured by the tactile sensor array in real time to estimate tangential forces from normal contact measurements. These tactile signals are utilized to penalize the operator’s aggressive operations, filter out unintentional hand movements, and as a result, stabilize the applied force in the manipulation. The VLA-based action refiner reasons over the input tactile images and the task context, and generates refined robot actions towards high precision in force control. The outputs from both the contact force stabilizer and the action refiner are fused and used to adjust the manipulation plan from the operator’s pure input, eventually driving the robot’s manipulation. Therefore, the major contributions of this work are summarized as follows:
\begin{itemize}
    \item We present XRoboToolkit-T, a high-frequency, tactile-enabled teleoperation system that incorporates tactile sensor signals for efficient data collection. It can be adapted to both gripper- and dexterous hand-based teleoperation in contact-rich manipulations involving deformable objects.
    \item We propose a contact force stabilizer and a VLA-based action refiner that effectively assist the operator in achieving stable and precise force control during contact-rich manipulations.
    \item We conduct thorough experiments on challenging tasks (including grasping rubber pipettes to transfer liquids, grasping plastic bottles and paper cups, and inserting medical syringe needles into soft materials) to verify the effectiveness and efficiency of XRoboToolkit-T.
\end{itemize}

\section{Related Works}

\subsection{Teleoperation}

With the increasing demand for precise manipulation in real-world applications, teleoperation systems have become a crucial tool for collecting expert demonstrations. In this context, the quality of collected data plays a critical role in enabling effective robot learning. Prior work has shown that teleoperated demonstrations can be used to acquire manipulation skills and significantly improve policy learning efficiency through high-quality demonstration data \cite{Si2021TeleopLfD}. In particular, applications with zero-tolerance requirements, such as robotic surgery\cite{Lee2018surgery}, have established standards for system latency and haptic feedback, highlighting the importance of precise and reliable teleoperation interfaces.
More recent studies further revealed that the modality of demonstration e.g., kinesthetic teaching, teleoperation, or vision-only observation, significantly affects imitation learning performance \cite{Li2025DemonstrationModality}.
Usability is another critical dimension of teleoperation systems. Recent vision-based frameworks, such as AnyTeleop \cite{Qin2023AnyTeleop}, have significantly lowered the barrier to entry by enabling sophisticated arm-hand coordination through standard camera observations and hand tracking, eliminating the need for specialized wearable hardware. Building on this accessibility, Open-Teach \cite{Iyer2025OpenTeach} provides a versatile platform specifically optimized for large-scale data acquisition to support robot learning. Furthermore, research into low-cost hardware configurations \cite{Zhao2023LowCostBimanual} demonstrates that even complex, fine-grained bimanual manipulation policies can be effectively learned from human demonstrations captured via accessible interfaces.

\subsection{XR Human Robot Interaction}
XR-enabled teleoperation allows operators to interact with robots in 3D space through spatial perception and embodied interaction \cite{Wang2024XRHRIReview}, providing a more immersive manipulation experience for users. Systems such as XRoboToolkit leverage depth-enhanced visual feedback to enable intuitive control of robotic manipulators, allowing operators to perceive spatial relationships and guide end-effector motions more effectively during teleoperation.
Immersive visual feedback systems further improve operator situational awareness and manipulation accuracy by providing active viewpoints and stereoscopic feedback \cite{Cheng2024OpenTelevision}. XR technologies have also been used to accelerate robot manipulation research in physics-based simulation environments. Physics simulators such as MuJoCo \cite{Todorov2012MuJoCo} provide OpenXR-based interfaces standardize cross-device interaction, and enable unified input from VR controllers and hand-tracking devices. In addition, platforms such as Isaac Lab \cite{mittal2025isaaclab} support XR-enabled visualization and teleoperation in simulation, allowing operators to control robots in the immersive environment \cite{wang2025ropesim}.

\subsection{Contact-rich Manipulation}
Contact-rich manipulation is crucial for applications that require precise control and stable physical interaction. Recent robot control models, such as $\pi_0$ \cite{Black2024PiZero} and OpenVLA \cite{Kim2024OpenVLA}, learn manipulation policies conditioned on visual observations and semantic instructions. However, the absence of tactile feedback in these approaches limits their ability to reason about contact dynamics and regulate interaction forces during manipulation. Subsequent work, such as $\pi_{0.6}^{*}$\cite{pi_star_0_6_2025}, further improves manipulation performance through reinforcement learning fine-tuning, demonstrating complex tasks such as robotic coffee preparation. Meanwhile, recent surveys \cite{Urain2024DeepGenerativeRobotics,Sapkota2025VLAReview, wang2026vlabot} highlight the potential of leveraging multimodal large-scale human demonstrations to bridge perception and action in robotic learning.
In tactile-enabled robot manipulation, several studies have demonstrated the importance of tactile sensing for contact-rich tasks. Prior works \cite{zhang2025vtla, hughes2020simple} have shown that tactile feedback can significantly improve performance in assembly tasks such as peg insertion. Similarly, \cite{yuan2024robotsynesthesia} leveraged visuo–tactile sensing to enable in-hand object rotation, while \cite{huang2024vitac} utilized tactile feedback to achieve reliable grasping of fragile objects.
Meanwhile, large-scale robotics initiatives have increasingly emphasized the importance of collecting real-world interaction data to bridge language reasoning and physical execution \cite{GeminiRobotics2025}.

\section{Teleoperation Architecture}

\subsection{Tactile Perception}

\begin{figure}
  \centering
  \includegraphics[width=\linewidth]{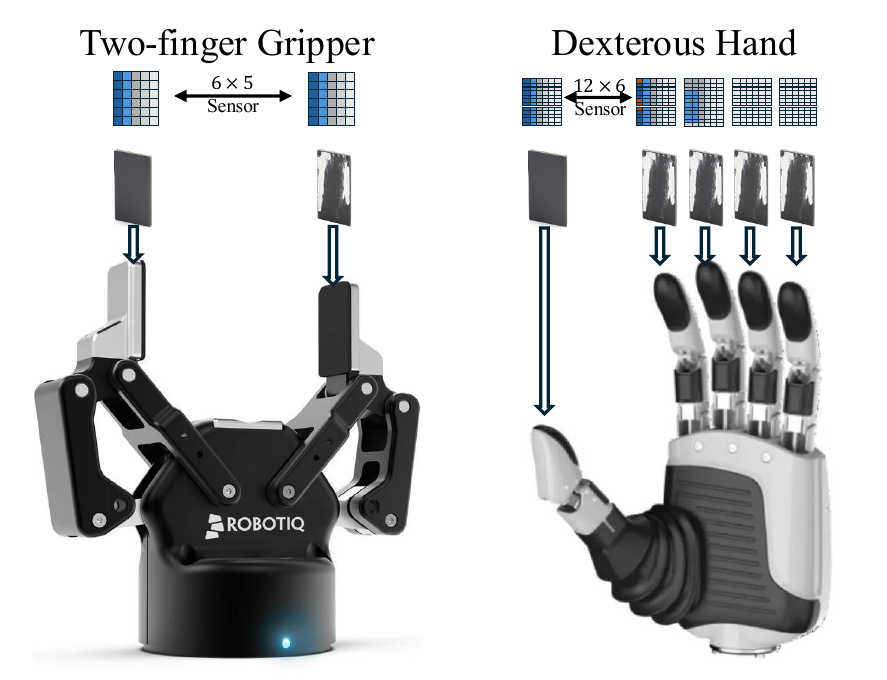}
  \caption{Tactile sensing for two-finger grippers and dexterous hands. The system supports both end-effectors with integrated tactile sensors and external third-party tactile sensing modules.
  }
  \label{endeffector}
\end{figure}

As shown in Figure \ref{endeffector}, we integrate two robotic end-effectors equipped with tactile sensing into the teleoperation system.
For devices without built-in tactile sensing, the first configuration employs a parallel two-finger Robotiq gripper augmented with third-party tactile sensors. The XRT-T system provides a generic sensor reading module to support the integration of external tactile devices. The recorded tactile data are represented as a pressure grid with a resolution of $x \times y$. The module defines a hardware-agnostic interface that abstracts vendor-specific communication protocols and converts raw measurements into a unified tactile data format compatible with the robot feedback stream. Incoming signals are synchronized with robot kinematics and timestamped to maintain temporal consistency across perception channels.
For devices with integrated tactile sensing, such as dexterous hands equipped with embedded tactile arrays (e.g., Realhand L6), the XRT-T system can also utilize the same generic sensor reading module for data acquisition. In this configuration, tactile perception data include pressure distributions, shear force measurements, combined force direction, and end-effector contact status.

Both sensing systems measure spatial pressure distributions across the contact surface, allowing the robot to infer contact geometry, force distribution, and object pose during grasping. In the visualization, gray regions indicate weak or no contact, while the heatmap represents the magnitude of the measured pressure, with higher intensity corresponding to greater contact force. The XRT-T  system extend the tactile sensing ability in XRoboToolkit, and can publish pressure distributions, force details, and derived contact features (e.g., contact state) to PICO XR devices or other display hardwares based on \textit{PC Service} package in XRoboToolkit \cite{zhao2025xrobotoolkit}.

\subsection{Dexterous Hand Control}

The XRT-T system supports a range of dexterous end-effectors with different actuation complexities and sensing capabilities. The control interface is designed to be morphology-agnostic, enabling the operation of sensorless dexterous hands, compact six-actuator hands, and high-degree-of-freedom hands within the same teleoperation pipeline. A unified command abstraction maps user inputs and teleoperation commands to joint-level actuation while internally handling differences in kinematic structure and joint limits.
The dexterous hand controller supports multiple command sources and can operate in coordination with different robot arm controllers and hand-tracking inputs.

Two primary input modalities are supported.
(1)The XRT-T system utilizes hand gesture tracking from the PICO 4 Ultra device and retargets the captured hand motion to the robotic dexterous hand. The human hand pose is represented by 26 joint poses, including four joints for the thumb and five joints for each of the remaining fingers. The retargeting module converts the tracked human hand configuration into the corresponding robot joint commands while respecting the mechanical constraints of the dexterous hand. (2)The XRT-T system supports controller input with a transmission frequency of up to 90 Hz. Users can define specific hand manipulation functions through configurable button mappings on the controller.

\begin{figure*}
  \centering
  \includegraphics[width=\linewidth]{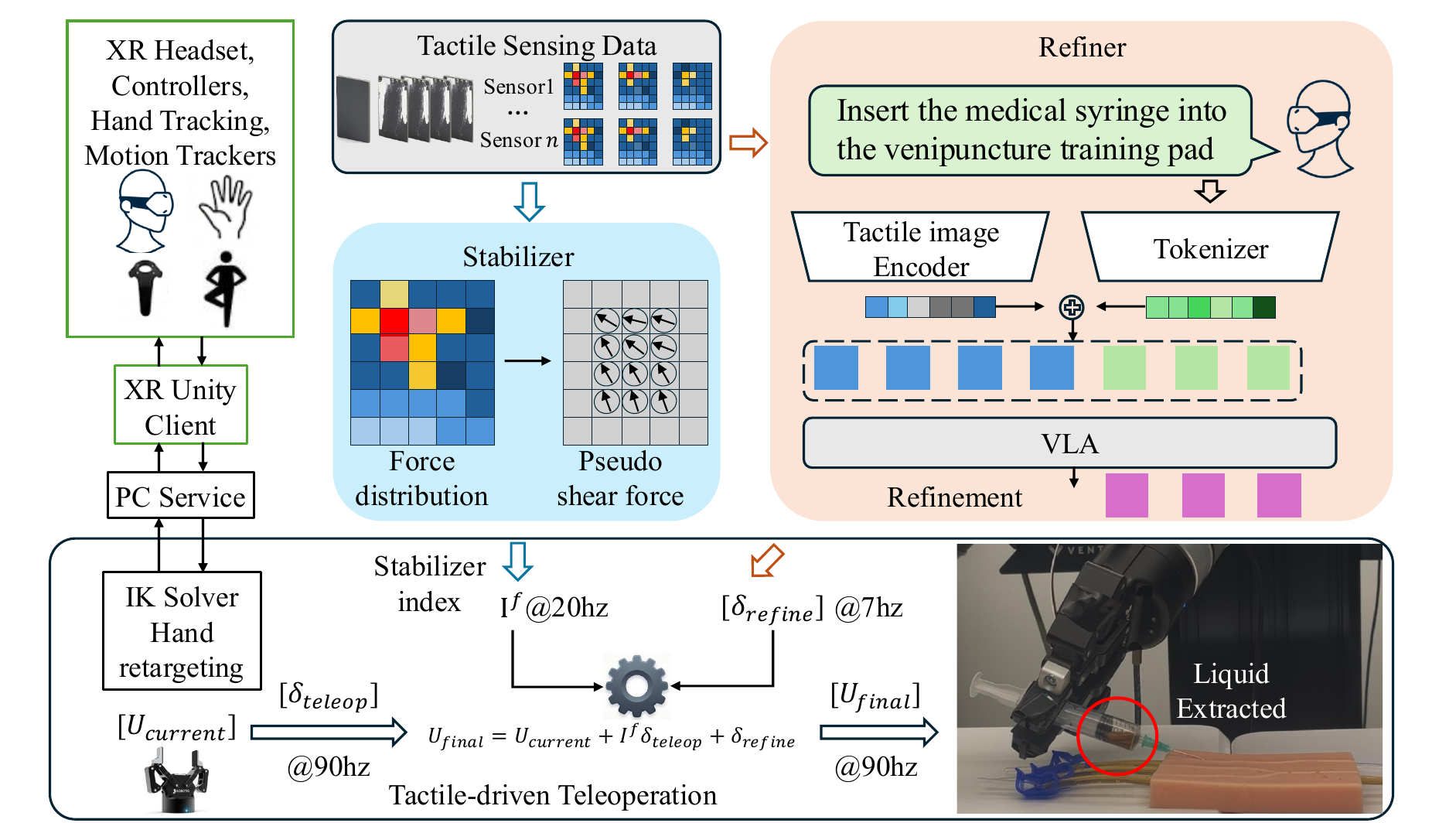}
  \caption{Overview of the XRoboToolkit-T teleoperation system. The framework integrates a tactile stabilizer and a tactile refiner to assist teleoperation, improving stability and enabling reliable data collection for contact-rich manipulation tasks.
  }
  \label{qualitative}
\end{figure*}

\begin{table}[t]
\caption{Robotic Haptic Feedback Data Formats}
\centering
\begin{tabular}{p{1.3cm} p{2.3cm} p{3cm}}
\hline
\textbf{Type} & \textbf{Field} & \textbf{Description} \\
\hline

\textbf{Tactile Perception} & Normal Force & Tactile normal pressure, each sensor have specific grid size ($x \times y$) \\
 & Shear Force &  Force that acts parallel to a surface \\
  & Force Direction & direction of combined force\\
 & Contact State & Contact status list of all sensor (0: none, 1: contact) \\

\hline
\textbf{End-Effector} & Type & Type of equippment on end effector\\
& Active Dof& number of active degree of freedom\\
& Pose & End-effector pose in base frame including position($p_x,p_y,p_z$) and orientation ($q_x,q_y,q_z,q_w$) \\
 & Velocity & End-effector linear/angular velocity \\

 & Torque & Force torque on the end effector \\
 & End Effector Status & Status of each Dof  \\
\hline
\textbf{Robotic Joints} & Mode & Index of different type of robots\\
& Joint Position & Joint angles for all Dofs \\
 & Joint Velocity & Joint velocities  \\
 & Joint Effort & Joint effort/torque feedback  \\
 & Chassis movement & Angular and linear speed of chassis movement\\
 & Position & Location of robot\\
\hline

\hline
\end{tabular}
\label{tab:robot_feedback}
\end{table}
\subsection{Teleoperation with haptic feedback}

The robot data feedback stream module extends the XRoboToolkit \textit{PC Service} to support real-time bidirectional communication between the robot system and client applications (Table \ref{tab:robot_feedback}). The module adopts an asynchronous publish–subscribe pipeline to continuously transmit multimodal robot feedback, including joint states, end-effector pose, and tactile/force sensing data. A unified data interface encapsulates sensor measurements into structured packets, while the \textit{PC Service} manages network transport, synchronization, and event dispatch to registered callbacks. This design enables low-latency monitoring and responsive interaction, allowing external applications to access robot state information and react to physical interactions during teleoperation and manipulation tasks.

\paragraph{Tactile Perception}
The tactile perception module provides high-resolution contact feedback for contact-rich manipulation tasks. The normal force represents the pressure distribution measured by each tactile array, where every sensor is organized in a predefined grid structure of size
$x\times y$.
The shear force captures tangential forces acting parallel to the contact surface, which are critical for detecting incipient slip and analyzing object stability during grasping or in-hand manipulation.
The contact state is represented as a binary list corresponding to all sensing elements  (0:
no contact, 1: contact), enabling fast detection of contact
transitions and supporting high-frequency feedback control.

\paragraph{End-effector}
The End-Effector module describes the structural configuration and dynamic state of the robotic manipulation.
The Type field defines the physical category of the end-effector. By explicitly specifying the end-effector type, the system can automatically adapt control strategies, perception pipelines, and data interfaces to match the hardware capabilities. This design allows users to flexibly integrate different grippers or dexterous hands within the same robotic framework.
The Active DOF (Degrees of Freedom) specifies the number of independently actuated joints in the end-effector. This parameter distinguishes between different manipulation devices, such as simple two-finger grippers, multi-finger grippers, or high-DOF anthropomorphic dexterous hands.
The Pose describes the spatial configuration of the end-effector in the robot base frame,including position($p_x,p_y,p_z$) and orientation ($q_x,q_y,q_z,q_w$). This enables precise spatial reasoning and coordination with perception modules.
The Velocity includes both linear and angular velocity of the end-effector, supporting dynamic motion control and trajectory tracking.
The Torque (force–torque wrench) represents the interaction forces and moments acting on the end-effector, which are critical for compliant control and contact-aware manipulation.
Finally, the End-Effector Status provides the state of each active DOF, such as  operational condition, allowing monitoring of joint-level behavior and fault detection.

\paragraph{Robot Joints}
The Robotic Joints module provides the global state representation of the robot platform, including joint-level feedback and mobile base information. This module enables unified monitoring and control across heterogeneous robotic embodiments.
The Mode defines the robot morphology and platform type.
The Joint Position contains the joint angles for all degrees of freedom.
The Joint Velocity represents the velocities of each joint.
The Joint Effort provides joint-level torque or effort feedback.
The Chassis Movement describes the motion state of the robot base, including linear velocity and angular velocity.
Finally, the Position indicates the global or local spatial location of the robot within a reference frame.

\section{Tactile-driven Control}

\subsection{Tactile assistance}
The tactile assistance module is composed of two components: a real-time stabilizer and a refiner.
The real-time stabilizer estimates a pseudo shear force from the temporal variation of the normal force distribution on the tactile array. Let $P_t(i,j)$ denote the normal pressure measured at sensor cell $(i,j)$ at time $t$. The maximum normal pressure is defined as
$
P_{\max}^{t} = \max_{(i,j)\in\Omega} P_t(i,j)
$.
The direction of the pseudo shear force is approximated by the displacement of the maximum-pressure location between consecutive frames
$\Delta \mathbf{c}^{t} = \mathbf{c}^{t} - \mathbf{c}^{t-1}$,
where $\mathbf{c}^{t}$ denotes the location of the maximum-pressure cell.
The magnitude of the pseudo shear force is proportional to the maximum normal pressure

\begin{equation}
s^{t} = k_s P_{\max}^{t}
\end{equation}

The coefficient $I^{f}$ acts as a stabilizing control factor that modulates the teleoperation command based on the estimated pseudo shear force. The coefficient is inversely related to the shear magnitude: when the shear force becomes large, $I^{f}$ approaches zero to suppress teleoperation commands; when the shear force remains below a predefined threshold, $I^{f}$ approaches one, allowing normal teleoperation control.

In addition, abrupt changes in the direction of the maximum-pressure location indicate unstable contact conditions. In such cases, the stabilizer temporarily disables teleoperation by setting $I^{f}=0$.

\begin{equation}
I^{f} =
\frac{1}{1 + \alpha \|\mathbf{F}_{ps}^{t}\|}, \|\mathbf{F}_{ps}^{t}\| \leq F_{\text{th}}
\end{equation}.
The stabilizer coefficient is designed as 20Hz frequency to ensure the stability and consistency of tactile feedback during robotic manipulation.

The tactile refiner is implemented as a Vision–Language–Action (VLA) model that performs high-level corrective adjustments during contact-rich manipulation. Instead of using conventional RGB visual input, the model takes a tactile image representing the spatial pressure distribution from the tactile sensor as its visual observation. The language input is a task-level instruction that describes the manipulation objective, such as “Insert the medical syringe into the venipuncture training pad.”
Given the tactile observation and the task description, the VLA model predicts an action vector representing the corrective motion $\delta_{\text{refine}}$ of the robot end-effector.
The refiner updates the corrective action
$\delta_{\text{refine}}$ at a lower frequency of approximately 7 Hz. The output corresponds to a small Cartesian displacement of the end-effector along with the
$x, y, z$ directions.

\subsection{Tactile-driven teleoperation}

To support stable contact-rich manipulation, the teleoperation system integrates tactile-driven assistance at multiple temporal scales. The tactile-driven teleoperation combines a stabilizer module and a refinement formulated as follows.
\begin{equation}
\mathbf{U}_{ee}^{t+1} =
\mathbf{U}_{ee}^{t}
+
I^{f}\,\boldsymbol{\delta}_{\text{teleop}}^{t}
+
I^{s}(t)\,\boldsymbol{\delta}_{\text{refine}}^{t}
\end{equation}
Let $\delta_{teleop}^t$
 denote the operator-intended displacement of the robot end-effector at time
$t$. The stabilizer, running at approximately 20 Hz, analyzes the spatial force distribution obtained from the tactile array. Based on the temporal variation of the pressure map, a pseudo shear-force index is computed to estimate the stability of the current contact state. This index modulates the operator command through a control coefficient
$I^f$
, which suppresses excessive motion when unstable shear forces or abrupt contact changes are detected.

In parallel, a VLA model processes tactile observations as structured tactile images and integrates them with task-level context to infer manipulation-aware adjustments. This module operates at a lower frequency and outputs a task-level motion refinement
$\delta_{refine}^t$, which compensates for small pose errors during contact-rich interactions.
The refiner only updates the end effector when a new refinement signal becomes available.This behavior is represented by a synchronization indicator $I^{s}(t)$. $I^{s}(t)=1$if the tactile refinement is updated at time $t$. Otherwise $I^{s}(t)=0$
to ensure that the refinement term is applied only at the update moments.

\begin{figure*}
  \centering
  \includegraphics[width=\linewidth]{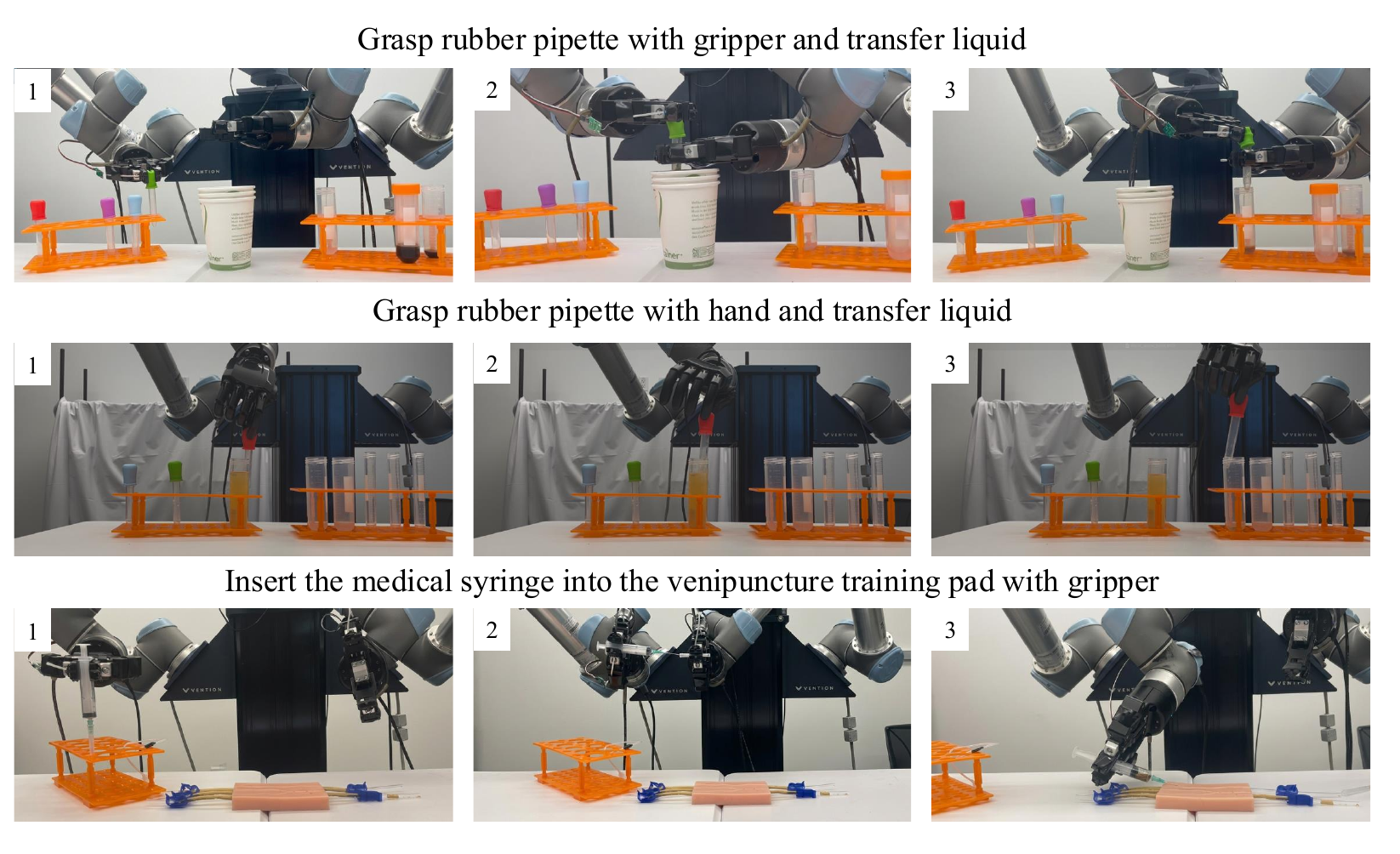}
  \caption{Contact-rich manipulation application overview. The medical syringe insertion and rubber pipette liquid transfer task require stable and precise teleoperation for different end-effectors.
  }
  \label{application_gripper}
\end{figure*}

\begin{figure}
  \centering
  \includegraphics[width=\linewidth]{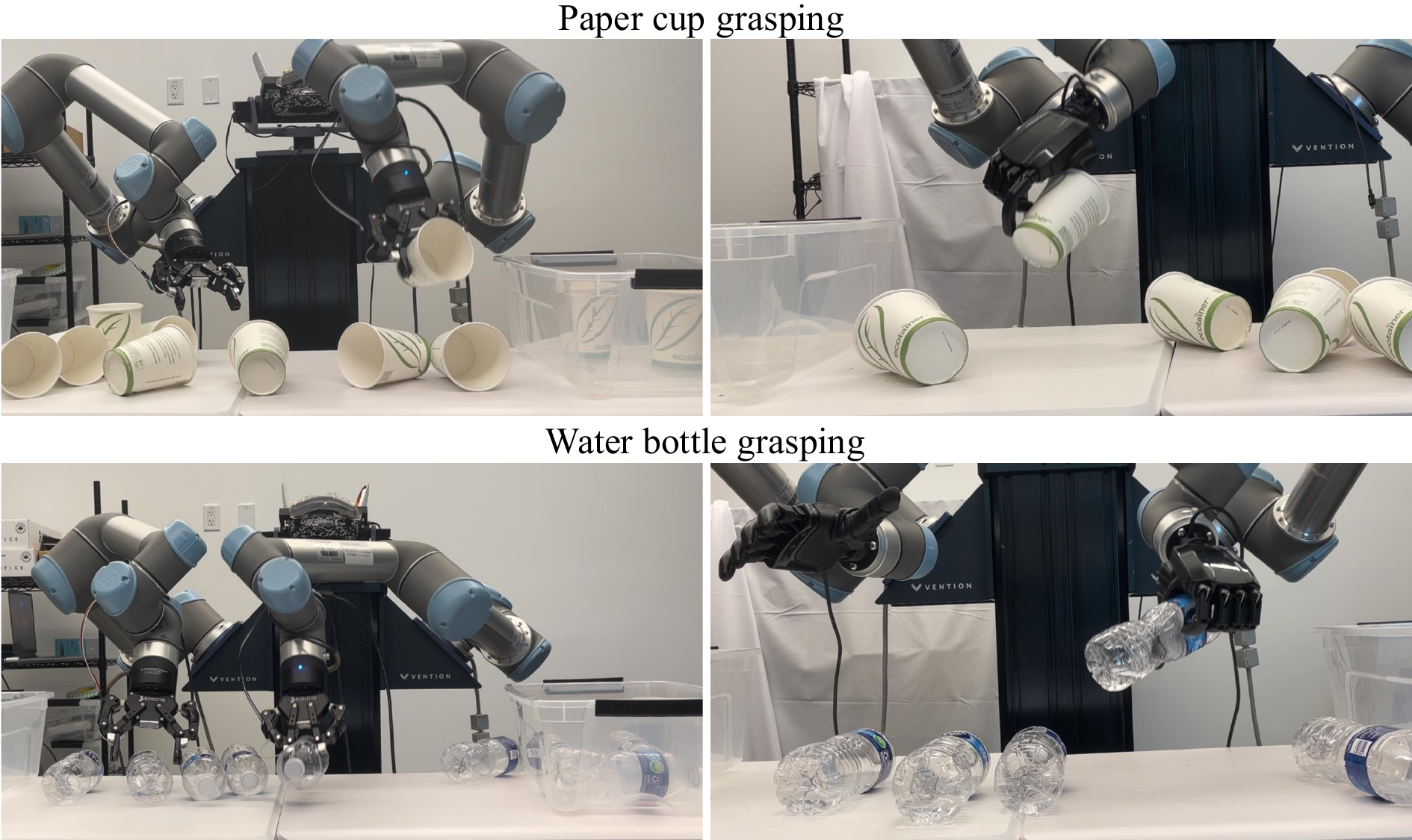}
  \caption{Dual arm grasping using both gripper and dexterous hand.
  }
  \label{grasping}
\end{figure}
\section{Experiments and Demonstrations}

Figure \ref{application_gripper} and Figure \ref{grasping} illustrates the applications and hardware configurations for different contact-rich manipulation tasks. In the two-finger gripper setup, two Robotiq grippers are mounted on the end-effectors of dual UR5e manipulators. Each gripper is equipped with two $6 \times 5$ piezoresistive tactile sensor arrays, enabling contact force measurement during manipulation. In dexterous hand setup, two Realhand L6 dexterous hands with 6 motors replace the gripper. Each dexterous hand integrates five $12 \times 6$ piezoresistive tactile sensor arrays.

\subsection{Two-finger grippers and dexterous hands application}

\paragraph{Medical Syringe}
Figure \ref{application_gripper} illustrates a liquid extraction task using a medical syringe as a representative contact-rich manipulation scenario. The complete procedure begins with the robot retrieving a syringe stored on an orange holder, where the syringe body is constrained inside a 1.6 cm diameter slot. After grasping the syringe, the robot removes the protective needle cap in free space. The cap has a length of 5.2 cm and a diameter of 8 mm, requiring stable grasping and coordinated pulling motion to detach it without disturbing the needle. The exposed needle is 4.5 cm long with a diameter smaller than 1 mm, making the subsequent manipulation highly sensitive to pose and force errors.
The robot then moves toward a venipuncture training pad that serves as a skin phantom for fluid extraction. The pad measures $11.1 cm \times 7 cm \times 1.5 cm$ and contains three embedded artificial vessels, each with a diameter of approximately 5 mm. The needle must accurately enter one of the artificial vessels before the robot pulls the syringe plunger to extract the liquid.

\paragraph{Rubber Pipette}
The second task involves transferring liquid using a rubber pipette and represents another contact-rich manipulation scenario requiring careful force regulation and stable grasping. The procedure begins with the robot retrieving the pipette from an orange holder, where it is stored in a 1.6 cm diameter slot. The pipette has a total length of 10.5 cm, consisting of a rigid tube with a diameter of 1.4 cm and a deformable rubber bulb measuring 3 cm in length and 2.1 cm in diameter.
The task is performed using both a two-finger gripper and a dexterous hand.

\paragraph{Grasping}
The dual-arm grasping application in Figure \ref{grasping} demonstrates coordinated manipulation with tactile feedback for handling multiple water bottles and paper cups. In this task, the robot is required to grasp each object from a cluttered workspace, lift it securely, and place it into a designated container.

\begin{table}[t]
\caption{Number of successful manipulations in 15 minutes for different end-effectors}
\centering
\begin{tabular}{lc}
\hline
\textbf{Manipulation Mode}
& \textbf{Manipulations in 15 minutes}  \\
\hline
Twist2 \cite{ze2025twist2} & 128 \\
XRT-T + Two-finger gripper &  138 \\
XRT-T + Dexterous hand      &  132  \\

\hline
\end{tabular}
\label{manipulation_efficiency}
\end{table}

\subsection{Results}

\paragraph{Manipulation efficiency}
In the experiment, we first evaluate the manipulation performance and data collection efficiency of the proposed teleoperation system with tactile feedback. The robotic system is tasked with grasping water bottles and paper cups and placing them into designated boxes. Detailed descriptions of the paper cup and water bottle experiments are provided in the Appendix.
Table~\ref{manipulation_efficiency} reports the number of successful manipulations completed in 15 minutes under different manipulation configurations. In the paper cup grasping experiment, XRT-T equipped with a two-finger gripper achieves the highest throughput, completing 138 successful grasps within 15 minutes, demonstrating its high efficiency in structured pick-and-place tasks. XRT-T with a dexterous hand and tactile sensing achieves 132 successful grasps in 15 minutes. In comparison, the state-of-the-art teleoperation system without tactile sensing (Twist2) completes 128 successful manipulations within the same duration.
These results indicate that the proposed teleoperation system with tactile sensing achieves superior data collection efficiency while maintaining competitive manipulation performance.

\begin{table}[t]
\caption{Dexterous Hand Control Frequency Benchmark}
\centering
\begin{tabular}{lcc}
\hline
\textbf{Approach (Control Mode)}
& \textbf{Mean Frequency (Hz)}
& \textbf{STD (Hz)  } \\
\hline

XRT \cite{zhao2025xrobotoolkit} + hand tracking & 20.2 & 3.5   \\
XRT-T + hand tracking      & 48.6 & 3.2   \\
XRT-T + controller           & 96.4 & 4.2   \\

\hline
\end{tabular}
\label{dex_frequency}
\end{table}

\paragraph{Control frequency}
In the experiment, we also test the control frequency of contact-rich manipulation in dexterous hand.
Table~\ref{dex_frequency} summarizes the control update frequency of the end-effector under different XRoboToolkit versions and control modalities. XRT represents the original teleoperation system \cite{zhao2025xrobotoolkit} and XRT-T depicts the proposed  teleoperation system with tactile feedback.
The original XRT system using hand tracking achieves a relatively low control rate (20.2 Hz), which is mainly limited by the perception update rate and communication latency. After upgrading to XRT-T with hand-tracking, the control frequency increases to 48.6 Hz, indicating improved system architecture and optimized data transmission.
When tactile sensing is enabled, the control frequency further increases to 96.4 Hz. This higher rate is possible because the tactile module provides direct contact feedback independent of the vision pipeline, allowing faster and more stable control updates during manipulation.
\begin{figure}
  \centering
  \includegraphics[width=\linewidth]{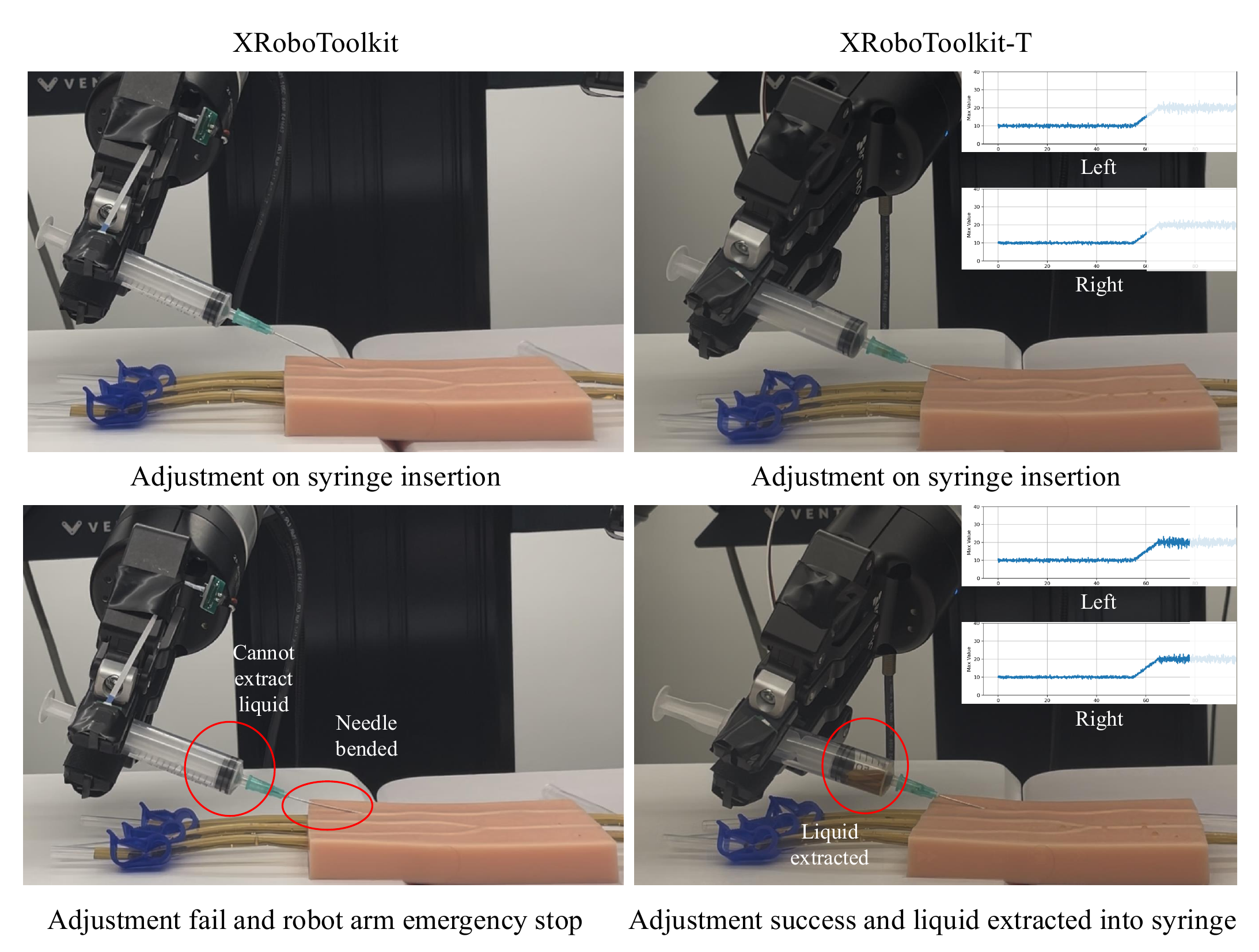}
  \caption{Qualitative comparison between XRoboToolkit and XRoboToolkit-T on the syringe insertion task.
Without tactile feedback (left), the needle bends during insertion.
With tactile-driven assistance (right), the system performs micro-adjustments during insertion and successfully extracts liquid.
  }
  \label{insertion}
\end{figure}
\paragraph{Qualitative results}
The qualitative performance of the proposed XRoboToolkit-T was evaluated against the baseline XRoboToolkit on the syringe needle insertion task using the same UR5 robot arm (Figure \ref{insertion}).
As illustrated in the experimental results, without tactile assistance the XRoboToolkit system struggled with the rigid contact constraints during the needle insertion process. As a result, the needle bent during the insertion attempt and the robot failed to extract liquid from the target. The accumulated contact force further generated excessive torque at the end-effector, which triggered the robot safety emergency stop and caused the task to fail.
In contrast, the tactile-driven assistance provided by XRoboToolkit-T enabled the robot to handle the significantly higher precision required for syringe manipulation. By continuously refine the end effector, the robot was able to maintain stable contact between the needle and the target surface, ultimately enabling successful liquid extraction.
These results demonstrate that tactile-driven teleoperation significantly improves the stability and precision of robotic manipulation in contact-rich tasks.

\subsection{Ablation study}

\begin{table}[t]
\caption{Number of successful manipulations per minute}
\centering
\begin{tabular}{lc}
\hline
\textbf{Manipulation Mode}
& \textbf{Manipulations per minute}  \\
\hline
XRT \cite{zhao2025xrobotoolkit} & 8.5 \\
XRT-T + refiner &  9.0 \\
XRT-T + stabilizer       &  8.8  \\

XRT-T + stabilizer + refiner & 9.6\\

\hline
\end{tabular}
\label{ablation}
\end{table}

We conduct an ablation study on the water bottle and paper cup grasping tasks to evaluate the impact of tactile sensing and adaptive motion modulation (tactile-informed force control module) on manipulation efficiency. The performance metric is defined as manipulations per minute, which reflects the data collection efficiency during repetitive grasp–transfer cycles.
The baseline method (XRT) operates without tactile sensing and relies primarily on visual perception and predefined grasp strategies, achieving 8.5 manipulations per minute. When tactile sensing is enabled (XRT-T + refiner), the system achieves 9.0 manipulations per minute, demonstrating that tactile feedback improves grasp stability and reduces recovery time caused by slip or misalignment.
Introducing a conservative motion profile (XRT-T + stabilizer) results in 8.8 manipulations per minute. This configuration enhances grasp reliability for deformable objects such as paper cups by preventing excessive compression.
Finally, combining adaptive tactile-informed force control modulation (XRT-T + stabilizer + refiner) yields the best performance at 9.6 manipulations per minute. In this configuration, the system dynamically switches between cautious contact phases and rapid transfer phases based on tactile feedback. This strategy reduces failure-induced delays while maintaining high transfer speed.

\section{CONCLUSIONS}
In this work, we presented XRoboToolKit-T, a tactile-driven teleoperation system designed to enable stable and precise assistance for contact-rich manipulation. By integrating high-resolution tactile sensing with a stabilizer and a refiner architecture, the system provides multi-timescale tactile assistance during teleoperation. The stabilizer haptic module analyzes pressure distributions from tactile arrays to estimate contact forces and support real-time operator assistance, while the refiner module leverages a VLA model to generate refinement based on tactile observations and task descriptions.
We validated the proposed system on several challenging contact-rich manipulation tasks, including rubber pipette grasping and liquid transfer, medical syringe insertion, and the manipulation of water bottles and paper cups. Experimental results demonstrate that the proposed tactile-driven teleoperation improves stability and increases the efficiency of collecting manipulation demonstrations.
In future work, we plan to extend the system to more complex industrial manipulation scenarios.

\section*{APPENDIX}
 The experiments were conducted on two different hardware configurations: a dual-UR arm platform equipped with dexterous hands and a dual-UR arm platform equipped with parallel grippers. The teleoperation system was controlled using a Pico4U XR headset, which communicated with the control computer via WiFi to stream operator motion commands to the robot system in real time. We also support the insprie hand RH56DFX for dexterous hand testing without a tactile sensor. End-effectors can be controlled by either controller trigger button or hand tracking in the XR headset. Haptic glove users may transform the hand to a universal format (MANO\cite{romero2022embodied} or mediapipe \cite{zhang2020mediapipe}) and control the dexterous hand.
For each configuration, the operator performed continuous manipulation trials involving grasping and placing deformable objects, including paper cups and water bottles. Each experimental condition was evaluated over 15 minutes of teleoperation, during which successful manipulation events were recorded and analyzed to measure the efficiency of contact-rich manipulation data collection.

\bibliographystyle{IEEEtran}

\bibliography{reference}

\end{document}